\documentclass[10pt,twocolumn,letterpaper]{article}

\usepackage[algorithms]{wacv}      

\usepackage{graphicx}
\usepackage{amssymb}
\usepackage{booktabs}
\usepackage{amsmath,amsfonts}
\usepackage{algpseudocode}
\usepackage{algorithm}
\usepackage{array}
\usepackage{cuted}

\usepackage{textcomp}
\usepackage{url}

\usepackage{epsfig}
\usepackage{subcaption}
\usepackage{multirow}
\usepackage{float}
\usepackage{breqn}
\usepackage{balance}

\usepackage[pagebackref,breaklinks,colorlinks]{hyperref}

\usepackage[capitalize]{cleveref}
\crefname{section}{Sec.}{Secs.}
\Crefname{section}{Section}{Sections}
\Crefname{table}{Table}{Tables}
\crefname{table}{Tab.}{Tabs.}

\def\wacvPaperID{*****} % *** Enter the WACV Paper ID here
\def\confName{WACV}
\def\confYear{2025}

\begin{document}

\title{InDistill: Information flow-preserving knowledge distillation \\ for model compression}

\author{Ioannis Sarridis$^1$ \qquad Christos Koutlis$^1$ \qquad Giorgos Kordopatis-Zilos$^2$ \\ Ioannis Kompatsiaris$^1$ \qquad Symeon Papadopoulos$^1$  \vspace{3pt} \\
$^1$Information Technologies Institute, CERTH\\
$^2$ VRG, FEE, Czech Technical University in Prague\\
% 6th km Charilaou-Thermi Rd, Thessaloniki, 57001, Greece\\
{\tt\small \{gsarridis,ckoutlis,ikom,papadop\}@iti.gr} ~~~~{\tt\small kordogeo@fel.cvut.cz}
}

\maketitle

\begin{abstract}
In this paper, we introduce InDistill, a method that serves as a warmup stage for enhancing Knowledge Distillation (KD) effectiveness. InDistill focuses on transferring critical information flow paths from a heavyweight teacher to a lightweight student. This is achieved via a training scheme based on curriculum learning that considers the distillation difficulty of each layer and the critical learning periods when the information flow paths are established. This procedure can lead to a student model that is better prepared to learn from the teacher.
To ensure the applicability of InDistill across a wide range of teacher-student pairs, we also incorporate a pruning operation when there is a discrepancy in the width of the teacher and student layers. This pruning operation reduces the width of the teacher's intermediate layers to match those of the student, allowing direct distillation without the need for an encoding stage.
The proposed method is extensively evaluated using various pairs of teacher-student architectures on CIFAR-10, CIFAR-100, and ImageNet datasets demonstrating that preserving the information flow paths consistently increases the performance of the baseline KD approaches on both classification and retrieval settings. 
The code is available at
\url{https://github.com/gsarridis/InDistill}.

\end{abstract}

%%%%%%%%% BODY TEXT
\section{Introduction}
\label{sec:intro}

\begin{figure}[t]
    \centering
    \includegraphics[width=\linewidth, trim={0cm 0cm 0cm 0cm},clip]{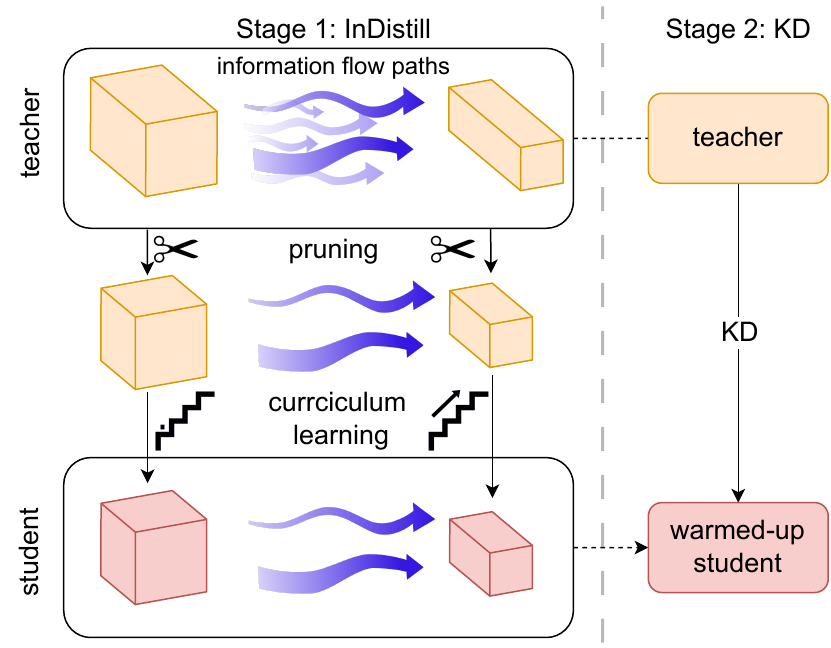}
    \vspace{-15pt}
    \caption{\textbf{Overview of the proposed InDistill method}. In stage 1, InDistill acts as a warmup, transferring the teacher's critical information flow via curriculum learning and direct distillation through pruning. This enhances the effectiveness of the main knowledge distillation process in the next stage 2.
    \vspace{-10pt}
    }
    \label{fig:teaser}
\end{figure}

The constant effort to increase the performance of Deep Neural Networks (DNN) has resulted in %much 
deeper architectures that require massive compute power and memory to train and deploy. Also, several application settings need to deploy DNNs on hardware with limited resources, which has increased the popularity of the research field of model compression. During the past years, model compression has attracted a lot of research interest and many pertinent approaches have been proposed such as parameter pruning, quantization, low-rank factorization, and Knowledge Distillation (KD) \cite{choudhary2020comprehensive, cheng2018model}.

The objective of KD, %which constitutes 
one of the most effective ways for model compression \cite{cheng2018model,gou2021knowledge}, is to transfer knowledge from a powerful %teacher 
network to a smaller and faster one, to extend its performance capabilities. Early KD approaches aim at transferring the teacher's logits \cite{hinton2015distilling}. This enables smaller models to understand how bigger models perceive the inputs, which enhances the learning of the former based on knowledge captured by the latter. However, the lack of intermediate layer supervision impedes the learning of how the information flows through the teacher's layers reducing the student's generalization potential \cite{passalis2020heterogeneous}. To remedy this, recent KD methods \cite{zagoruyko2016paying,romero2014fitnets} distill knowledge from intermediate layers, as it is empirically shown to improve the KD effectiveness. \looseness=-1

Pruning methods are applied to reduce storage requirements or inference time \cite{choudhary2020comprehensive, liang2021pruning, ma2019resnet}. Unstructured pruning \cite{yang2021comparative} removes unimportant weights reducing the model's storage size, while structured pruning \cite{li2016pruning,luo2017thinet, ding2021resrep, chen2020dynamical, wang2021convolutional} removes the less important CNN filters reducing both the model's storage size and processing time. Furthermore, prior work have explored combining KD with pruning for model compression. Some approaches apply pruning before KD~\cite{park2022prune}, while others prune and subsequently apply KD only to the layers unaffected by pruning~\cite{aghli2021combining} or use KD to fine-tune pruned models~\cite{chen2018shallowing}. 
However, none of the aforementioned methods follow the standard teacher-student paradigm; thus, they cannot be compared with typical KD approaches. \looseness=-1  

Furthermore, it is worth noting that during the training process, a neural network undergoes several phases \cite{achille2017critical}. More precisely, the first training epochs are considered responsible for the creation of the model's information flow paths. 
These paths represent the routes through which information travels and is processed within the network, effectively shaping the network's ability to learn. It is important to highlight that after their formation, these paths can only be fine-tuned through the rest of the training procedure.
This critical fact has been taken into consideration in prior work~\cite{passalis2020heterogeneous} to effectively distill intermediate layers, without, however, taking into consideration the increasing distillation difficulty of successive intermediate layers, from shallower to deeper network parts.

This paper introduces InDistill, a method that addresses the aforementioned limitations and can be employed as a warmup stage prior to any KD approach (see \cref{fig:teaser}).
Inspired by curriculum learning principles \cite{bengio2009learning} and the critical learning periods of a neural network \cite{passalis2020heterogeneous}, InDistill suggests a simple, yet effective way to distill the teacher's information flow paths to the student, by directly transferring knowledge from each intermediate layer separately and in ascending transferring difficulty order (\ie, from shallow to deep layers). 
Additionally, to expand the applicability of InDistill, channel pruning is specifically applied to the teacher's intermediate layers when a width disparity exists between the teacher and student architectures.
This selective pruning aligns the architectures, facilitating direct feature map matching and thus preserving the crucial information flow paths. Similarly, in scenarios with an extremely high capacity gap between the teacher and student models, an auxiliary teacher is also considered to facilitate the application of InDistill.

The proposed method has been evaluated on a wide range of classification and retrieval settings with the use of widely adopted benchmarks (\ie, CIFAR-10 \cite{krizhevsky2009learning}, CIFAR-100 \cite{krizhevsky2009learning},  and ImageNet \cite{deng2009imagenet}) combined with various state-of-the-art KD approaches and is found to boost their performance in the vast majority of our conducted experiments.
The main contributions of this work are:
\vspace{-4pt}
\begin{itemize}
    \setlength\itemsep{-4pt} 
    \item A method that introduces a curriculum learning-based training scheme considering the distillation difficulty of each layer and the critical learning periods crucial for effectively transferring the teacher's information flow paths to the student, as a warmup step before employing any KD approach.
    \item A pruning mechanism applied in scenarios where there is a width disparity between teacher and student architectures. This mechanism aligns the architectures, facilitating direct knowledge transfer from feature maps and preserving the crucial information flow paths.
    \item A wide analysis involving 5 baseline KD methods and 3 widely-adopted benchmarks, demonstrating the effectiveness of InDistill in improving the performance of the baseline approaches.
\end{itemize}

\section{Related work}
\label{sec:related}
\textbf{Logits-based knowledge distillation.} 
KD, initially proposed in \cite{hinton2015distilling}, is an effective way for model compression by reproducing the teacher's class probability distribution using a much faster and smaller student model. Several KD methods exploiting only the model's output probability distributions have been proposed \cite{kim2017transferring, muller2019does, meng2019conditional, zhu2018knowledge, park2021learning, deng2021comprehensive}.
A recent work \cite{zhao2022decoupled} reformulates the classical KD loss into target class knowledge distillation and non-target class knowledge distillation, demonstrating the efficacy of logit distillation with this Decoupled Knowledge Distillation (DKD) methodology. 
The Multi-Level KD (MLKD) \cite{jin2023multi} approach enhances logit distillation through multilevel prediction alignment, by incorporating batch-level and class-level prediction alignments. Furthermore, logit standardization \cite{sun2024logit} prior to the distillation can address the mismatch in logit magnitudes between teacher and student models and thus improve the distillation effectiveness. 

\textbf{Features-based knowledge distillation.} The first method that leverages feature maps of the teacher in order to provide the student with extra supervision is presented in Romero \etal~\cite{romero2014fitnets}. This method transfers knowledge from only one intermediate layer with the aid of feature map encoding. Similarly, Probabilistic Knowledge Transfer (PKT) \cite{passalis2018learning} distills knowledge from the penultimate layer's feature maps by trying to match the teacher's and student's distributions of the samples' pairwise similarities. However, transferring knowledge of one intermediate layer can not capture the critical connections between the layers. To alleviate this shortcoming, the Attention Transfer (AT) \cite{zagoruyko2016paying} proposes an attention mechanism that applies distillation to all intermediate layer representations. Similarly, the Hierarchical Self-supervised Augmented Knowledge Distillation (HSAKD) \cite{yang2021hierarchical} employs classifiers on top of all intermediate layers to supervise the KD process. 
In a different perspective, ReviewKD \cite{chen2021distilling} introduces cross-stage connection paths in KD, emphasizing the importance of inter-level connections between teacher and student networks. 
Furthermore, Tian \etal~\cite{tian2019contrastive} introduces Contrastive Representation Distillation (CRD) which uses a contrastive loss to distill the feature maps that derive from the last convolutional layer.
Kim \etal~\cite{kim2018paraphrasing} introduces a method to effectively encode the extracted features before KD, but still encoding is necessary. The aforementioned methods share the same shortcomings, they disregard the preservation of information flow paths during distillation, ignore the capacity gap between the models (\ie, the difference in the number of learnable parameters), and face the challenge of transferring knowledge from multiple layers simultaneously.

\textbf{Information flow preservation.} An effort to capture the information flow was made in Yim \etal~\cite{yim2017gift} by generating a Flow of Solution Procedure (FSP) matrix that captures the relation between two successive layers. Additionally, Passalis \etal~\cite{passalis2020heterogeneous} consider a loss function based on mutual information in order to transfer the information flow paths. Also, it introduces a critical-periods-aware Weight Decay (WD) scheme that reduces the learning rate of the feature-based KD after each epoch, motivated by the fact that the first training epochs are responsible for creating the information flow paths \cite{achille2017critical}.

\textbf{Auxiliary teacher.}
Mirzadeh \etal~\cite{mirzadeh2020improved} propose the usage of an auxiliary model in order to reduce the capacity gap between teacher and student models. It should be stressed though, that they do not utilize the intermediate layers. Based on the same idea, Passalis \etal~\cite{passalis2020heterogeneous} makes use of an auxiliary model to alleviate the structural differences between teacher and student.

\textbf{Curriculum learning} Several fields have utilized curriculum learning approaches \cite{graves2017automated, soviany2022curriculum, gong2016multi, khan2011humans, zhang2021curriculum}, which suggest splitting a hard task into sub-tasks and learning them sequentially based on difficulty order \cite{bengio2009curriculum, zaremba2014learning}. In Matiisen \etal~\cite{matiisen2019teacher}, a teacher-student curriculum learning framework is introduced for reinforcement learning, where the teacher determines the sub-tasks that the student should be trained on at each training step, while Pentina \etal~\cite{pentina2015curriculum} proposes learning tasks sequentially for enhancing multi-task learning effectiveness. However, none of the existing curriculum learning approaches considers the model's layers as sub-tasks of increasing difficulty or is designed to retain the informational flow paths. 

\textbf{Knowledge distillation and pruning.} Finally, there have been a few attempts \cite{aghli2021combining,chen2018shallowing} to combine KD and pruning \cite{li2016pruning} for model compression. However, neither the mixing techniques nor the approaches' objectives coincide with ours. More specifically, Aghli \etal~\cite{aghli2021combining} prunes some layers of a model and distills the rest, while Chen \etal~\cite{chen2018shallowing} applies KD on a pruned model. Instead, we prune the teacher model in order to have an equal width with the student and straightforwardly transfer knowledge from intermediate layers. 

\section{Methodology}
\label{sec:method}

\subsection{Problem formulation}
\label{subsec:problem}
The problem of transferring knowledge from a teacher to a student model is formulated as follows. Let $\mathbf{X}\in\mathbb{R}^{3\times h\times w}$ denote an input image with $h$ and $w$ height and width, respectively. $d(\cdot)$ is the teacher model, and $l=1,\dots,L_d$ its layer indices. Then, $\mathbf{P}^{(l)} = d(\mathbf{X},l) \in \mathbb{R}^{n_{d,l}\times h_{d,l} \times w_{d,l}}$ denotes the teacher's $l$-th layer feature map, where $n_{d,l}$, $h_{d,l}$, and $w_{d,l}$ is its number of channels and spatial dimensions. Accordingly, consider a student model $g(\cdot)$ with $\theta$ learnable parameters,
$L_g$ layers and $\mathbf{S}^{(l)} = g(\mathbf{X},l) \in \mathbb{R}^{n_{g,l}\times h_{g,l} \times w_{g,l}}$ the $l$-th layer's feature map. In addition, let $\mathbf{q}_t$, $\mathbf{q}_s$ $\in\mathbb{R}^C$ be the class probability distributions of the teacher and student model, respectively, with $C$ the number of output classes. Then, the goal of a typical logit-based KD is to match the teacher's and student's class probability distributions ($\mathbf{q}_t\simeq\mathbf{q}_s$). In contrast, InDistill aims to prepare the student model $g(\cdot)$ for this process by learning $\theta$ parameters that can enhance KD effectiveness.

\subsection{Information flow preservation}
\label{sec:info_preserv}
First, let us consider the simplest scenario, where teacher and student architectures share the same width, and thus it is feasible to transfer knowledge from the corresponding feature maps directly. Specifically, the loss between $l$-th layer feature maps, $\mathbf{P}^{(l)}$ and $\mathbf{S}^{(l)}$ is defined as:
% \looseness=-1
\begin{equation}
\label{eq:loss2}
    \mathcal{L}_{MSE}^{(l)} = \| \mathbf{P}^{(l)} - \mathbf{S}^{(l)} \|_2^2, %\ l \in \{1,2,\cdots,L_{g-1}\}
\end{equation}
where $\| \cdot \|_2$ denotes the $l_2$-norm. Note that InDistill can only be applied to intermediate layers.

\begin{figure}[t]
    \centering
    \includegraphics[width=0.75\linewidth, trim={0 0cm 0 0cm},clip]{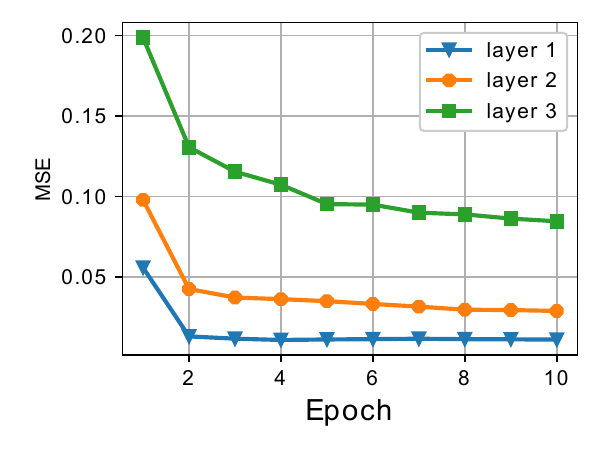}
    \caption{\textbf{$\mathcal{L}_{MSE}$ loss curves.} KD is applied from a teacher to a 3-layer student using PKT~\cite{passalis2018learning} loss on the CIFAR-10 dataset. The MSE loss is computed based on the feature maps of images from the validation set.
    }
    \label{fig:curr_just}
\end{figure}

Curriculum learning suggests dividing a hard task into sub-tasks w.r.t. their difficulty and performing training sequentially in increasing difficulty. When trying to transfer useful information from all the teacher's layers, one could hypothesize that transferring knowledge from shallow layers is easier than deep layers, for two reasons. First, shallow layers hold in general low-level information (\eg, edges and corners), while deep layers hold task-specific high-level information. Second, deep layer distillation also requires transferring the knowledge obtained by all previous layers as well as the information flow paths until then. \Cref{fig:curr_just} exemplifies the higher loss when transferring knowledge from deeper layers, validating this intuition empirically.
Given the above, we propose a curriculum learning scheme that considers the distillation of each layer as a separate sub-task. Particularly,  $L_g$ sequential sub-tasks are spawned equal to the number of student layers. These sub-tasks determine the training schedule. If the total number of training epochs is $E$, then the number of epochs corresponding to each sub-task (\ie, layer) can be calculated as: 
\begin{equation}\looseness=-1
\label{eq:ep}
    e_i = \left\{
    \begin{matrix}
        a + i b & i\neq L_g\\ 
        E -\sum_{i=1}^{L_g-1} e_i & i = L_g
    \end{matrix}
    \right., i \in \{1,2,\cdots,L_g\},
\end{equation}
where $a$ is a threshold parameter indicating the minimum number of epochs for each layer's training and $b$ is a parameter that increments the number of epochs w.r.t. the sub-task's difficulty. Also, the set of epochs that corresponds to each sub-task $i$ is $\mathcal{S}_i = \{r_{i}+1, r_{i}+2, \cdots, r_i+e_i\}$ where:\looseness=-1
\begin{equation}
    r_i = \left\{
    \begin{matrix}
        0 & i=1\\ 
        \sum_{k=1}^{i-1}e_{k} & i > 1
    \end{matrix}
    \right ., i \in \{1,2,\cdots,L_g\}.
\end{equation}
Note that in each sub-task $i$, the first $i$ layers are trained. By adopting the proposed curriculum learning scheme, the first $\sum_{i=1}^{L_g-1} e_i$ epochs are dedicated to the preservation of the teacher's information flow paths. After this warmup stage, any existing KD loss can be employed, combined with a task-specific loss (\eg, the cross-entropy loss for the classification tasks). The final loss is calculated as:
\begin{equation}
\label{eq:curr}
    \mathcal{L} = \left\{
    \begin{matrix}
     \vspace{2pt}
        \mathcal{L}_{MSE}^{(i)} & i\neq L_g\\
        \mathcal{L}_{KD}  + \mathcal{L}_{task}& i = L_g
    \end{matrix}
    \right ., i \in \{1,2,\cdots,L_g\},
\end{equation}
This way, the student model can effectively form the critical connections that facilitate the following KD process. 

\subsection{Channel pruning}

\label{subsec:prune}
 
\begin{figure}[t]
    \centering
    \includegraphics[width=\linewidth]{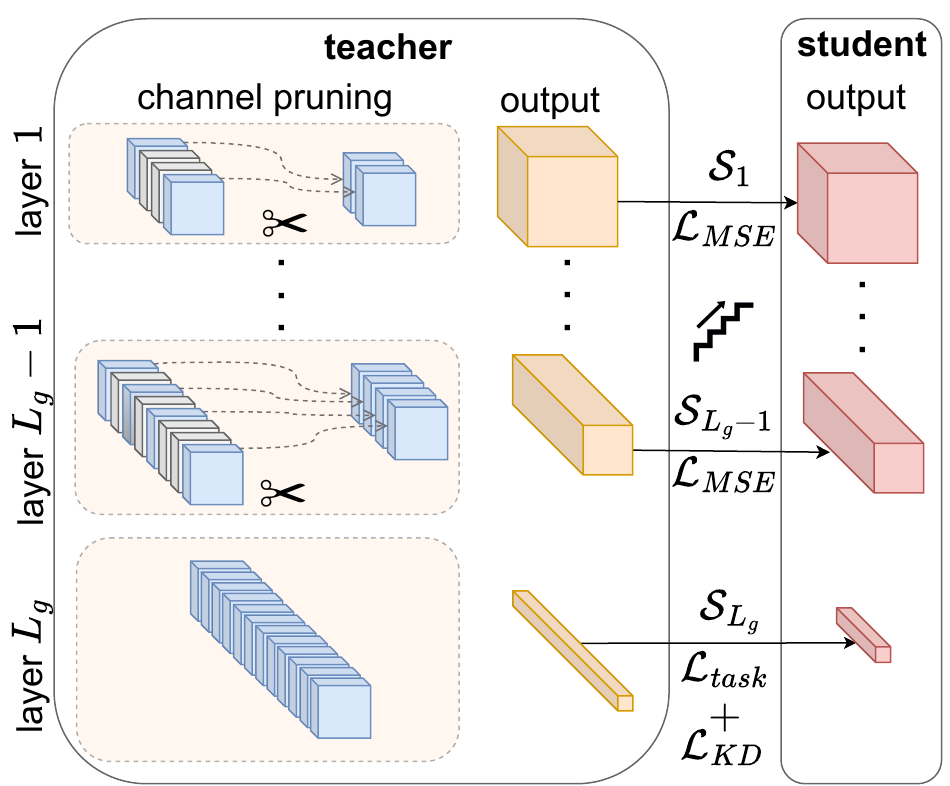}
    \caption{
    \textbf{Channel pruning and curriculum learning visualization}. In cases of large capacity gap, channel pruning is applied on the teacher's intermediate layers so that teacher and student $l \in (1, L_g-1)$ layer widths coincide. Then, curriculum learning is applied, where each $l$ layer is trained for $\mathcal{S}_l$ epochs.
    After applying InDistill, any knowledge distillation loss can be employed as $\mathcal{L}_{KD}$ along with the $\mathcal{L}_{task}$ task loss.
    }
    \label{fig:arch}
\end{figure}

In more complex KD scenarios, where there is a width discrepancy between the teacher and student networks, channel pruning is employed to force architectural alignment and thus allow for direct KD on feature maps (\ie, $\mathcal{L}_{MSE}$).
The typical criterion for evaluating the importance of a filter is the  $l_1$-norm or $l_2$-norm \cite{wen2016learning, han2015learning, li2016pruning}. Here, we opt for the approach that applies structured channel pruning based on $l_1$-norm~\cite{li2016pruning}. Specifically, 
let $\mathbf{K} \in \mathbb{R}^{n_i \times n_o \times k \times k}$ denote the kernel of an arbitrary layer, where $k$ is the kernel size, and $n_i$ and $n_o$ the input and output channels, respectively. Then, we follow the pruning procedure described in \cref{alg:prune}.
\begin{algorithm}[t]
\caption{Channel pruning procedure}
\label{alg:prune}
\begin{algorithmic}[1]
    \State \textbf{Inputs}: filters $\mathbf{K}$ and the number of filters to prune $p$. \textbf{Output}: the pruned filters $\mathbf{K'}\in \mathbb{R}^{n_i \times (n_o-p) \times k \times k}$.
    \Procedure{Prune}{$\mathbf{K}$, $p$}
    \For{$i = 0$ \textbf{to} $n_o$}
    \State $\mathbf{s}_i\gets \sum_{j=1}^{n_i} \sum_{l=1}^k \sum_{m=1}^k \lvert \mathbf{K}_{j,i,l,m} \rvert$
    \EndFor
    \State $\mathbf{s}, \mathbf{x} \gets \textit{sort}(\mathbf{s})$ \Comment{$\mathbf{x}$ denotes the sorted indices array}
    \State $\mathbf{x} \gets \mathbf{x} \setminus \mathbf{x}[:p] $ \Comment{remove the first $p$ indices}
    \State $\mathbf{K'} \gets \mathbf{K}[:, \mathbf{x}, :, :]$  \Comment{get the $n_0 - p$ selected filters}
    % \State $\mathbf{K'} \gets \mathbf{K}[:, \mathbf{x}, :, :]$ \Comment{S remaining filters by indices}

    \State \textbf{return} $\mathbf{K'}$
    \EndProcedure
\end{algorithmic}
\end{algorithm}
Channel pruning is applied to all teacher's intermediate layers to force architectural alignment. After applying \cref{alg:prune} with $p=q\cdot n_{d,l}$, the teacher's pruned feature maps $\mathbf{P'}^{(l)} \in \mathbb{R}^{n_{g,l}\times h_{g,l} \times w_{g,l}}$ are of equal size to the student's feature maps, which allows for direct knowledge transfer involving no encoding. \cref{fig:arch} provides an overview of InDistill, including its pruning component.

\subsection{Auxiliary teacher}
In cases of teacher-student pairs with an extreme capacity gap, different number of layers $L_d \neq L_g$, or structurally different architectures (\eg, a teacher with and a student without residual connections), we
follow prior works~\cite{mirzadeh2020improved, passalis2020heterogeneous} that make use of an auxiliary model built from the teacher model using standard knowledge distillation schemes. $f(\cdot)$ denotes the auxiliary model, $L_f$ the number of layers (here $L_f$=$L_g$), and  $\mathbf{A}^{(l)} = f(\mathbf{X},l) \in \mathbb{R}^{n_{f,l}\times h_{f,l} \times w_{f,l}}$ its output feature maps. Note that $h_{f,l}$=$h_{g,l}$ and $w_{f,l}$=$w_{g,l}$ as the networks share kernel sizes.
The auxiliary teacher is designed to have the same number of layers and the same number of output channels as the student, after applying pruning with pruning rate $q$ to all intermediate layers. As a result, the auxiliary model's feature map sizes $n_{f,l}$ depend on the student model's feature maps sizes $n_{g,l}$ and the pruning rate $q\in[0,1)$, namely $n_{f,l} = \frac{n_{g,l}}{1-q}$. Having designed the auxiliary model, standard logit-based KD \cite{hinton2015distilling} is applied to transfer knowledge from the teacher to the auxiliary model.
\section{Experimental setup}
\label{sec:exp}

\subsection{Datasets and model architectures}
\label{sec:data}
The proposed method has been evaluated on three well-known datasets: CIFAR-10 \cite{krizhevsky2009learning},  CIFAR-100 \cite{krizhevsky2009learning}, and ImageNet \cite{deng2009imagenet}.
Regarding the model architectures, there are three types of setups based on the capacity gap and architectural differences between teachers and students:

\textbf{Small capacity gap (same width).} Following prior work~\cite{tian2019contrastive,zhao2022decoupled,jin2023multi}, we evaluate the proposed method using seven ResNet~\cite{he2016deep} teacher-student pairs widely used for evaluating KD methodologies (\eg, ResNet34-ResNet18, ResNet56-ResNet20, ResNet110-ResNet32, etc.). These pairs have the same width, meaning the feature maps derived from their blocks have matching dimensions. 

\textbf{Large capacity gap (different width).} We employ the following teacher-student pairs: ResNet50-ResNet18, ResNet32x4-ResNet32, ResNet32x2-ResNet20, and ResNet32x4-ResNet20. The teachers of these pairs have either 2 or 4 times the width of the corresponding students. In this scenario, channel pruning is applied to the teacher's intermediate layers to match the student's width, facilitating direct KD via feature-based distillation while preserving the crucial information flow paths. 

\textbf{Extreme capacity gap (different architecture).} Following prior work~\cite{passalis2020heterogeneous}, we consider ResNet18 as the teacher model (consisting of around 11 million trainable parameters) and a tiny CNN consisting of 3 convolutional layers as the student model (CNN-S). Due to the structural differences as well as the large capacity gap between teacher and student, we additionally use an auxiliary teacher model (CNN-A). \Cref{tab:aux_stud} presents the model architectures in detail. For CNN-A and CNN-S, batch normalization is applied after each convolutional layer. The feature maps are flattened before propagation to the fully-connected layers.

\begin{table}[t]
    \centering
    \resizebox{\linewidth}{!}{
    \begin{tabular}{c|c|c|c|c|c}
   
        model & \multicolumn{3}{c|}{\texttt{conv} (\#filters, \#kernel) } &  \multicolumn{2}{c}{\texttt{fc} (\#neurons)} \\
          \hline
        CNN-A & (16, 3$\times$3) & (32, 3$\times$3) & (64, 3$\times$3) & 128 & \#classes \\
        CNN-S & (8, 3$\times$3) & (16, 3$\times$3) & (32, 3$\times$3) & 64 & \#classes \\ \hline
    \end{tabular}
    }
    \caption{\textbf{Auxiliary teacher and student model architectures}. These network architectures are used for extreme capacity gap experiments on CIFAR-10. \texttt{conv} and \texttt{fc} stands for convolutional and fully-connected layer, respectively. }
    \label{tab:aux_stud}
\end{table}

\subsection{Implementation details and evaluation protocol}
\label{sec:train_det}

For the CIFAR-10 dataset, the models are trained using the Adam~\cite{kingma2014adam} optimizer for 60 epochs with a learning rate of 0.001 and for 10 epochs with a learning rate of 0.0001. The batch size is set to 128 and the parameters $a$ and $b$ are set to 2 and 1, respectively. For CIFAR-100, the models are trained for 240 epochs using the SGD~\cite{sutskever2013importance} optimizer with 0.9 momentum and 0.05 learning rate for the first 150 epochs, then the learning rate decays every 30 epochs by a factor of 10. The batch size is equal to 64 and the weight decay is 0.0005. For ImageNet, the models are trained for 100 epochs using the SGD optimizer with 0.9 momentum and 0.1 learning rate, while the learning rate decays every 30 epochs by a factor of 10. The batch size is equal to 256, and the weight decay is 0.0001. The parameters $a$ and $b$, are set to 5 and 1, respectively.

The official training/test splits are utilized for evaluation.  For this evaluation, we report accuracy (top-1 and top-5) for classification tasks, and mean Average Precision (mAP) and Precision@k (P@k) for retrieval tasks.
In the context of retrieval evaluation, the training sets are employed to create the databases. Subsequently, the test sets are used to query these databases, allowing us to measure the retrieval performance of representations.
Finally, we also evaluate the methods w.r.t. information flow preservation. The information flow of a network can be modeled through Mutual Information (MI) \cite{passalis2020heterogeneous}. MI divergence loss proposed in PKT\cite{passalis2018learning} aims at minimizing the MI divergence between teacher and student, thus forcing the student model to mimic the teacher's information flow paths. Hence, we consider it as a proper measure of information flow preservation and denote it as $\mathcal{L}_{MI}$.
The experiments were conducted using NVIDIA RTX-3060 and RTX-4090 GPUs.

\begin{table*}[t]
\centering

\resizebox{0.83\linewidth}{!}{
\begin{tabular}{lcccccc}
\toprule
\textbf{Teacher arch.} & \textbf{WRN-40-2} & \textbf{WRN-40-2} & \textbf{ResNet56} & \textbf{ResNet110} & \textbf{ResNet110} & \textbf{ResNet32x4}\\
\textbf{Student arch.} & \textbf{WRN-40-1} & \textbf{WRN-16-2} & \textbf{ResNet20} & \textbf{ResNet32} & \textbf{ResNet20} & \textbf{ResNet8x4}\\
\midrule
Teacher & 75.61 & 75.61 & 72.34 & 74.31 & 74.31 & 79.42\\
Student &71.98 & 73.26 & 69.06 & 71.14 & 69.06 & 72.50\\
\midrule
KD \cite{hinton2015distilling}& 73.54 & 74.92 & 70.66 & 73.08 & 70.67 & 73.33 \\
KD+InDistill & 75.09 \textcolor{blue}{(+1.55)} & 76.17 \textcolor{blue}{(+1.25)} & 72.16 \textcolor{blue}{(+1.50)} & 74.78 \textcolor{blue}{(+1.70)} & 72.29 \textcolor{blue}{(+1.62)} & 75.73 \textcolor{blue}{(+2.40)} \\
\midrule
PKT \cite{passalis2018learning}& 73.45 & 74.54 & 70.34 & 72.61 & 70.25 & 73.64 \\
PKT+InDistill &73.28 \textcolor{red}{(-0.17)} &74.59 \textcolor{blue}{(+0.05)} &70.36 \textcolor{blue}{(+0.02)} &72.92 \textcolor{blue}{(+0.31)} &70.38 \textcolor{blue}{(+0.13)} & 73.92 \textcolor{blue}{(+0.28)} \\
\midrule
CRD \cite{tian2019contrastive}& 74.14 & 75.48 & 71.16 & 73.48 & 71.46 & 75.51 \\
CRD+InDistill &74.22 \textcolor{blue}{(+0.08)} & 75.54 \textcolor{blue}{(+0.06)} & 71.79 \textcolor{blue}{(+0.63)} & 73.95 \textcolor{blue}{(+0.47)} & 71.85 \textcolor{blue}{(+0.39)} & 75.73 \textcolor{blue}{(+0.22)} \\
\midrule
DKD \cite{zhao2022decoupled}& 74.81 & 76.24 & 71.97 & 74.11 & 71.06 & 76.32 \\
DKD+InDistill & 75.39 \textcolor{blue}{(+0.58)} & 76.12 \textcolor{red}{(-0.12)} & 71.95 \textcolor{red}{(-0.02)} & 74.59 \textcolor{blue}{(+0.48)} & 71.98 \textcolor{blue}{(+0.92)} & 76.46 \textcolor{blue}{(+0.14)}\\
\midrule
MLKD \cite{jin2023multi}& 75.35 & 76.63 & 72.19 & 74.11 & 71.89 & 77.08 \\
MLKD+InDistill& 75.71 \textcolor{blue}{(+0.36)} & 76.92 \textcolor{blue}{(+0.29)} & 72.64 \textcolor{blue}{(+0.45)} & 74.68 \textcolor{blue}{(+0.57)} & 72.06 \textcolor{blue}{(+0.17)} & 76.99 \textcolor{red}{(-0.09)}\\
\bottomrule
\end{tabular}}
\vspace{-7pt}
\caption{\textbf{Evaluation on small teacher-student capacity gap.} Top-1 accuracy on CIFAR-100 of seven teacher-student homogeneous pairs trained with five KD methods with and without InDistill warmup. Performance of teachers and students without distillation is provided for reference. Numbers in parenthesis indicate performance \textcolor{blue}{gain} or \textcolor{red}{loss} from the use of InDistill.
\vspace{-13pt}
}
\label{tab:cifar100}
\end{table*}
\begin{table}[t]
\centering
% \small

\resizebox{0.77\linewidth}{!}{
\begin{tabular}{lcc}
\toprule
\textbf{method} & \textbf{top-1 acc} & \textbf{top-5 acc} \\ \midrule
Teacher & 73.31 & 91.42   \\
Student & 69.75 & 89.07 \\
\midrule
KD \cite{hinton2015distilling}& 71.03 & 90.05 \\
KD+InDistill & 71.34 \textcolor{blue}{(+0.31)} & 90.42 \textcolor{blue}{(+0.37)} \\ \midrule
 PKT \cite{passalis2018learning}& 70.41 & 89.48 \\
 PKT+InDistill & 71.13 \textcolor{blue}{(+0.72)} & 90.24 \textcolor{blue}{(+0.76)}\\ \midrule
 
 CRD \cite{tian2019contrastive}&  71.17 & 90.13 \\ 
 CRD+InDistill & 71.24 \textcolor{blue}{(+0.07)} & 90.37 \textcolor{blue}{(+0.24)}  \\  \midrule
 
 DKD \cite{zhao2022decoupled}& 71.70 & 90.41  \\ 
 DKD+InDistill &  71.87 \textcolor{blue}{(+0.17)} & 90.65 \textcolor{blue}{(+0.24)}  \\  \midrule

 MLKD \cite{jin2023multi}& 71.90 & 90.55  \\ 
 MLKD+InDistill & 72.03 \textcolor{blue}{(+0.13)} & 90.77 \textcolor{blue}{(+0.22)} \\  
 \bottomrule
\end{tabular}}
\vspace{-7pt}
\caption{\textbf{Large-scale evaluation on small capacity gap.} Top-1 and top-5 accuracy on ImageNet of five KD methods both with and without InDistill warmup. ResNet34 and ResNet18 are used as teacher and student networks.
\vspace{-8pt}
}
\label{tab:imagenet}
\end{table}
\subsection{Baseline methods}

To evaluate InDistill's effectiveness in enhancing the performance of baseline KD approaches, we consider 5 state-of-the-art methodologies: the original KD method (KD) \cite{hinton2015distilling}, the Probabilistic Knowledge Transfer (PKT) \cite{passalis2018learning}, the Contrastive Representation Distillation (CRD) \cite{tian2019contrastive}), the Multilevel Knowledge Distillation (MLKD) \cite{jin2023multi}, and the Decoupled Knowledge Distillation (DKD) \cite{zhao2022decoupled}.
Additionally, for comparison with the proposed scheme, we include in our evaluation two methods that claim to preserve the teacher's information flow: the Flow of Solution Procedure (FSP) \cite{yim2017gift} and the PKT-H-CR \cite{passalis2020heterogeneous}. 
More details regarding the aforementioned methods can be found in \cref{sec:related}. 

\section{Results}
\label{sec:res}
\subsection{Evaluation on standard KD benchmarks}
First, we evaluate InDistill on the CIFAR-100 dataset, employing the standard teacher-student pairs used for benchmarking of KD methodologies. The results are summarized in \cref{tab:cifar100}. For each teacher-student pair, we report the performance of the baselines compared with the performance of the baselines combined with the proposed InDistill warmup. InDistill consistently improves the performance across most methods, demonstrating significant gains. Notably, KD+InDistill shows a remarkable improvement in all teacher-student settings, with the highest gain of +2.40\% when WRN-32x4 is the teacher and WRN-8x4 is the student. Similarly, for CRD, the gain achieved by adding InDistill is notable, especially with a +0.63\% improvement for ResNet56 to ResNet20.

\begin{table}[t]
\centering
% \small
\resizebox{\linewidth}{!}{
\begin{tabular}{clccc}
\toprule
& \textbf{method} & \textbf{mAP} & \textbf{P@100} & \textbf{top-1 acc} \\ \midrule
% \multicolumn{4}{c}{ImageNet} \ \hline
\parbox[t]{2mm}{\multirow{8}{*}{\rotatebox[origin=c]{90}{\textbf{ImageNet}}}} & Teacher & 48.21 & 52.54 & 76.16  \\
&  Student & 24.79 & 25.13 & 69.75  \\
\cmidrule(lr){2-5}%\hline
& KD \cite{hinton2015distilling}& 30.91 & 31.98 & 71.33 \\
& KD+InDistill & 31.52 \textcolor{blue}{(+0.61)} & 32.53 \textcolor{blue}{(+0.55)} & 71.54 \textcolor{blue}{(+0.21)} \\ \cmidrule(lr){2-5}%\hline
% & PKT-H-CR & 21.92 & 23.86 & 68.86 \\
& PKT \cite{passalis2018learning}& 24.67 & 26.41 & 70.66 \\
& PKT+InDistill & 25.75 \textcolor{blue}{(+1.08)} & 27.46 \textcolor{blue}{(+1.05)} & 70.98 \textcolor{blue}{(+0.32)} \\ \cmidrule(lr){2-5}%\hline
& CRD \cite{tian2019contrastive}& 25.99 & 27.46 & 70.49 \\
& CRD+InDistill & 31.10 \textcolor{blue}{(+5.11)} & 32.19 \textcolor{blue}{(+4.73)} & 71.36 \textcolor{blue}{(+0.87)} \\ \midrule
% \multicolumn{4}{c}{CIFAR-100} \ \hline
\parbox[t]{2mm}{\multirow{8}{*}{\rotatebox[origin=c]{90}{\textbf{CIFAR-100}}}}& Teacher &  77.94 & 78.19 & 79.42  \\
&  Student &  43.42 & 57.42 & 71.14  \\
\cmidrule(lr){2-5}%\hline
& KD \cite{hinton2015distilling}& 56.59 & 65.08 & 72.11 \\
& KD+InDistill & 59.79 \textcolor{blue}{(+3.20)} & 67.45 \textcolor{blue}{(+2.37)} & 72.96 \textcolor{blue}{(+0.85)} \\ \cmidrule(lr){2-5}%\hline
% & PKT-H-CR & 46.40 & 58.52 & 70.36 \\
& PKT \cite{passalis2018learning}& 53.20 & 64.01 & 72.68 \\
& PKT+InDistill & 54.54 \textcolor{blue}{(+1.34)} & 64.78 \textcolor{blue}{(+0.77)} & 72.77 \textcolor{blue}{(+0.09)} \\ \cmidrule(lr){2-5}%\hline
& CRD \cite{tian2019contrastive}& 51.34 & 63.19 & 73.70 \\
& CRD+InDistill & 58.40 \textcolor{blue}{(+7.06)} & 67.98 \textcolor{blue}{(+4.79)} & 73.97 \textcolor{blue}{(+0.27)} \\ \bottomrule
\end{tabular} }
% \vspace{-7pt}
\caption{\textbf{Evaluation on large teacher-student capacity gap.} Classification and retrieval performance evaluation on ImageNet and CIFAR-100. ResNet50 and ResNet18 are used as teacher and student for the ImageNet dataset. ResNet32$\times$4 and ResNet32 are used as teacher and student for the CIFAR-100 dataset.
% \vspace{-10pt}
}
\label{tab:imagenet_cifar100}
\end{table}
\begin{table}[t]
\centering

\resizebox{\linewidth}{!}{
\begin{tabular}{lcccc}
\toprule
\textbf{Teacher arch.}& \textbf{ResNet32} & \textbf{ResNet32x2} & \textbf{ResNet32x4}  & \multirow{ 2}{*}{\textbf{std}} \\
\textbf{Student arch.}& \textbf{ResNet20} & \textbf{ResNet20} & \textbf{ResNet20} &  \\
\midrule
Teacher & 71.75 & 75.97 &  79.42&  - \\
Student & 69.06&  69.06&  69.06&  - \\
\midrule
KD \cite{hinton2015distilling}& 70.86& 69.84& 68.08&  1.41 \\
KD+InDistill & 71.99 \textcolor{blue}{(+1.13)} & 71.65 \textcolor{blue}{(+1.81)} & 70.96 \textcolor{blue}{(+2.88)} &  0.52 \textcolor{blue}{(-0.89)} \\
\midrule
PKT \cite{passalis2018learning}& 71.12& 69.94& 68.66&  1.23 \\
PKT+InDistill & 71.80 \textcolor{blue}{(+0.68)} & 71.92 \textcolor{blue}{(+1.98)} & 71.26 \textcolor{blue}{(+2.60)} &  0.35 \textcolor{blue}{(-0.88)} \\
\midrule
CRD \cite{tian2019contrastive}& 71.41& 70.62& 69.72&  0.85 \\
CRD+InDistill & 71.91 \textcolor{blue}{(+0.50)} & 72.24 \textcolor{blue}{(+1.62)} & 71.14 \textcolor{blue}{(+1.42)} & 0.56 \textcolor{blue}{(-0.29)} \\
\midrule
DKD \cite{zhao2022decoupled}& 71.16& 70.22& 68.47&  1.37 \\
DKD+InDistill & 71.94 \textcolor{blue}{(+0.78)} & 71.95 \textcolor{blue}{(+1.73)} & 71.11 \textcolor{blue}{(+2.64)} & 0.48 \textcolor{blue}{(-0.89)} \\
\midrule
MLKD \cite{jin2023multi}& 71.60& 71.99& 70.54 & 0.75 \\
MLKD+InDistill& 71.97 \textcolor{blue}{(+0.37)} & 72.48 \textcolor{blue}{(+0.49)} & 71.22 \textcolor{blue}{(+0.68)} &  0.63 \textcolor{blue}{(-0.12)} \\
\bottomrule
\end{tabular} }
\vspace{-7pt}
\caption{\textbf{Evaluation with increasing capacity gap.} Top-1 accuracy on CIFAR-100 for three network pairs. Std stands for the standard deviation of the top-1 accuracy over the three runs.
\vspace{-10pt}
}
\label{tab:cifar100_large}
\end{table}

\Cref{tab:imagenet} presents the results on the ImageNet dataset. The performance metrics include top-1 and top-5 accuracy for different methods, with and without the InDistill approach. The addition of InDistill yields improvements across all methods. For instance, KD+InDistill shows an increase of +0.31\% in top-1 accuracy and +0.37\% in top-5 accuracy. Similarly, for the top-performing method MLKD, MLKD+InDistill achieves gains of +0.13\% in top-1 accuracy and +0.22\% in top-5 accuracy, demonstrating the robustness and efficacy of the InDistill approach.

\subsection{Evaluation on large capacity gap settings}

Apart from the standard KD evaluation setups, we consider an additional teacher-student pair for CIFAR-100 and ImageNet datasets -- specifically, ResNet50-ResNet18 and ResNet32x4-ResNet32, respectively. In that way, we assess the contribution of pruning in our method while the capacity (and the width) of the teacher increases. The corresponding results are presented in \cref{tab:imagenet_cifar100}, concerning both classification and retrieval tasks. In terms of accuracy, the improvements are similar to the ones in small capacity gap experiments with gains ranging from +0.09\% to +0.87\%. Notably, for the retrieval task, we observe significantly higher improvements in terms of mAP and P@100\footnote{The presented values are computed based on cosine similarity.
}, such as +7.06\% mAP and +4.79\% P@100 for CRD+InDistill on CIFAR100. These gains can be attributed to the preservation of information flow paths, enabling better feature learning.
\looseness=-1

While the capacity gap between teacher and student increases, such as in the model pairs ResNet32-ResNet20, ResNet32x2-ResNet20, and ResNet32x4-ResNet20, traditional KD methods often suffer from diminished performance, as evidenced by \cref{tab:cifar100_large}. In particular, the baseline approaches generally result in decreased top-1 accuracy across most model pairs when width discrepancy increases. Integrating InDistill before applying KD significantly mitigates this issue. Specifically, it enhances the student's performance notably in the ResNet32x2-ResNet20 pair compared to the wider ResNet32-ResNet20 pair in all experiments, except for KD+InDistill, and demonstrates less accuracy drop from ResNet32x2-ResNet20 to ResNet32x4-ResNet20 pairs compared to baseline approaches. This trend is reflected in the standard deviation (std) values presented in \cref{tab:cifar100_large}, where InDistill consistently demonstrates reduced std compared to the baseline approaches (ranging from -0.12 to -0.89).
Furthermore, we investigate the effect of pruning all layers of the teacher model rather than only the specific layer being transferred. For the ResNet32x4-ResNet20 pair using KD as the primary knowledge distillation approach, we observe a 0.55\% accuracy drop (resulting in 70.41\% accuracy) compared to InDistill. This reduction occurs because pruning all layers drops the teacher model's overall capacity. Additionally, apart from width discrepancies, it is important to note that when spatial dimensions differ between teacher and student, pooling can be used to adjust them accordingly.
\begin{table}[t]
\centering

\resizebox{0.92\linewidth}{!}{
\begin{tabular}{lccc}
\toprule
\textbf{method} & \textbf{mAP} & \textbf{P@100} & \textbf{top-1 acc} \\

\midrule
Teacher & 90.47 & 92.26 & 95.23 \\
Auxiliary & 66.78 & 75.91 & 82.60\\
Student & 39.00 & 58.77 & 69.06 \\
\midrule
FSP  \cite{yim2017gift}& 37.66  & 57.41 & 71.14\\ 
PKT-H-CR \cite{passalis2020heterogeneous}& 53.00  & 64.06 & 73.33 \\
\midrule
KD \cite{hinton2015distilling}& 39.38 & 58.77 & 73.08 \\
KD+InDistill & 42.82 \textcolor{blue}{(+3.44)} & 61.09 \textcolor{blue}{(+2.32)} & 75.17 \textcolor{blue}{(+2.09)} \\
\midrule
PKT \cite{passalis2018learning}& 51.41 & 62.56 & 73.27 \\
% PKT-H-CR & 53.00  & 64.06 &  \\ 

PKT+InDistill & 54.61 \textcolor{blue}{(+3.20)} & 65.40 \textcolor{blue}{(+2.84)} & 74.13 \textcolor{blue}{(+0.86)} \\
\midrule
CRD \cite{tian2019contrastive}& 47.11 & 61.64 & 71.99 \\
CRD+InDistill & 49.57 \textcolor{blue}{(+2.46)} & 62.23 \textcolor{blue}{(+0.59)} & 73.32 \textcolor{blue}{(+1.33)} \\
\midrule
DKD \cite{zhao2022decoupled}& 42.82 & 61.11 & 74.45\\
DKD+InDistill & 44.08 \textcolor{blue}{(+1.26)} & 62.18 \textcolor{blue}{(+1.07)} & 74.84 \textcolor{blue}{(+0.39)}\\
\midrule
MLKD \cite{jin2023multi}& 43.24 & 61.54 & 73.76 \\
MLKD+InDistill & 45.40 \textcolor{blue}{(+2.16)} & 63.15 \textcolor{blue}{(+1.61)} & 74.87 \textcolor{blue}{(+1.11)}\\
\bottomrule
\end{tabular} }
\vspace{-7pt}
\caption{\textbf{Evaluation on extreme teacher-student capacity gap.}  Classification and retrieval performance evaluation on CIFAR-10. ResNet-18, CNN-A, and CNN-S are used as teacher, auxiliary, and student networks, respectively.
\vspace{-10pt}
}
\label{tab:cifar10}
\end{table}

\subsection{Evaluation on extreme capacity gap settings}

Finally, we evaluate the performance of the proposed method in extreme scenarios where there is a need to employ a very lightweight model (\eg, simple CNNs with only a few thousand parameters, as in \cref{tab:aux_stud}). In such cases, an auxiliary teacher is employed to allow for applying InDistill, due to the significant architectural differences between teacher and student models. 
\Cref{tab:cifar10} presents the performance of InDistill combined with 5 baseline methods.
Teacher and student baselines refer to from-scratch training without KD, while auxiliary baseline refers to performance after KD from the teacher. The proposed method consistently boosts the performance of all the core KD approaches in both retrieval and classification settings. Specifically, for the mAP metric, InDistill boosts performance significantly across methodologies, such as +3.44\% for KD, +3.20\% for PKT, +2.46\% for CRD, +1.26\% for DKD, and +2.16\% for MLKD.
Similarly, for the classification setting, InDistill increases the accuracy of all methods, with improvements ranging from +0.86\% to +2.09\%. Finally, in terms of latency the students' mean inference time is 14.89$\times$ faster than the teachers' (\ie, 0.49ms on CPU) and in terms of storage requirements the students' storage requirement is 28KB while teachers' is 42.7MB.

\begin{figure} [t]
    \centering
    \includegraphics[width=0.81\linewidth, trim=0.45cm 0.37cm .3cm 0.37cm, clip]{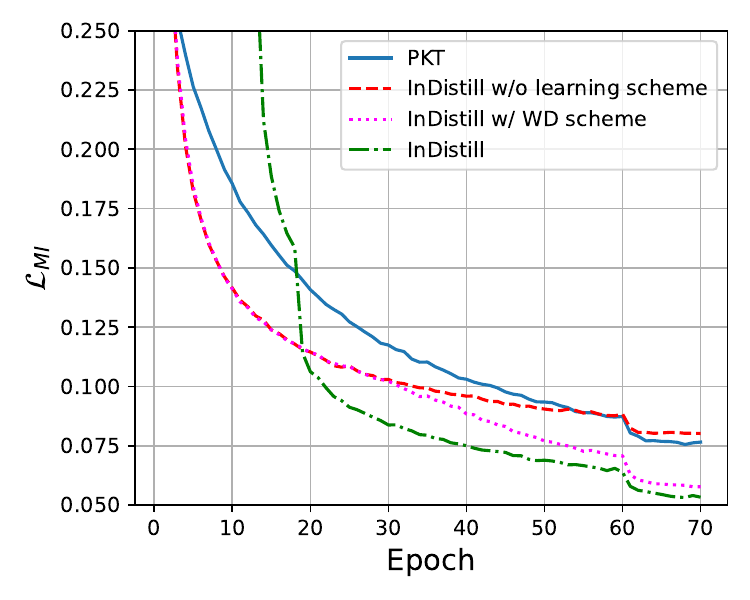}
    \vspace{-7pt}
    \caption{\textbf{Evaluation on information flow preservation.} $\mathcal{L}_{MI}$ loss curves during KD on CIFAR-10. Comparison of InDistill with the proposed, WD~\cite{passalis2020heterogeneous},  or without curriculum learning. PKT is used as the main KD approach.
    }
    \label{fig:abl}
\end{figure}

\subsection{Preserving information flow}
In \cref{fig:abl}, we present the $\mathcal{L}_{MI}$ loss curves to assess the proposed method's capability to distill the teacher's information flow. 
In particular, the loss curves of InDistill without a learning scheme (\ie, transferring knowledge from all layers simultaneously), with the WD scheme \cite{passalis2020heterogeneous}, and with the curriculum learning scheme, are depicted for comparison. Although InDistill without learning scheme reduces $\mathcal{L}_{MI}$ during the first training epochs (when the critical connections are created) compared to the baseline PKT method, it impedes the learning process during the last training epochs and achieves 51.54\% mAP. On the contrary, applying a learning scheme aware of the critical learning periods, such as WD and the proposed learning scheme, can successfully address this limitation. However, the proposed scheme is superior to WD, as not only is aware of the critical learning periods but it also takes into account the complexity of transferring the knowledge of multiple layers. This can also be confirmed by the mAP values being 54.61\% and 53.91\% for the proposed scheme and WD, respectively. Thus, we confirm that forcing the architectural alignment and being aware of the layers' successive transfer difficulty enables the student model to mimic the teacher's information flow more effectively. 
\looseness=-1
\subsection{Ablation study}
\begin{table}[t]
    \centering
    \begin{tabular}{ccc}
    \toprule
         \textbf{intermediate KD} & \textbf{learning scheme} & \textbf{top-1 acc}   \\
         \midrule
           - & - & 69.94  \\ 
           Eq.~\eqref{eq:loss2} & - & 68.64  \\ 
           PKT \cite{passalis2018learning}& WD \cite{passalis2020heterogeneous}&  70.02 \\
           PKT \cite{passalis2018learning}& Eq.~\eqref{eq:curr} &  71.06  \\ 
           Eq.~\eqref{eq:loss2} & Eq.~\eqref{eq:curr} &\textbf{71.92}  \\
         \bottomrule
    \end{tabular}
        \vspace{-7pt}
        \caption{
        \textbf{Impact of different components.}
         Top-1 accuracy on CIFAR-100. ResNet32x2 and ResNet20 are used as teacher and student. Comparison of the proposed intermediate distillation and curriculum scheme with PKT~\cite{passalis2018learning} and WD~\cite{passalis2020heterogeneous}, respectively. PKT is used as the main KD approach.
    }
    \label{tab:components}
\end{table}

\Cref{tab:components} presents how InDistill's components (\ie, teacher pruning and curriculum learning) cumulatively affect the student's performance. 
Furthermore, as already demonstrated in \cref{fig:abl}, pruning contributes to the critical connections' creation during the first training epochs, but eventually hinders the training procedure.
On the contrary, pruning combined with curriculum learning enables the critical connections formation as well as the effective learning of the final task, and results in considerably enhanced student's performance. 
\begin{figure}
    \centering
    \includegraphics[width=1.\linewidth, trim=1.5cm 0.0cm 1.5cm 1.3cm, clip]{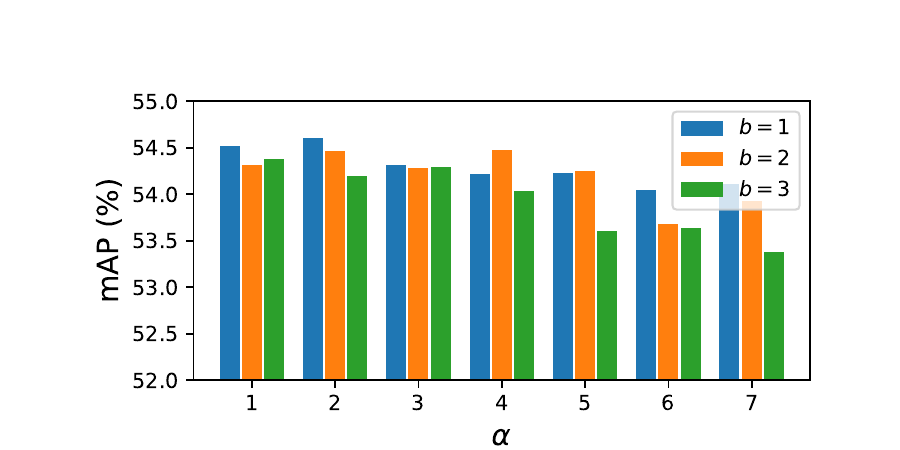}
        \vspace{-15pt}
        \caption{\textbf{Impact of curriculum parameters.} Student performance for different values of $a$ and $b$ on CIFAR-10. PKT is used as the main KD approach.
        \vspace{-5pt}
        }
        \label{fig:ab_ht}
\end{figure}
Finally, \cref{fig:ab_ht} presents how the hyperparameters $a$ and $b$ affect the student's performance when applying InDistill via our curriculum scheme.

\section{Conclusions}
\label{sec:conc}
In this work, we introduce a curriculum learning-based training scheme that acts as a warmup step to preserve the teacher model's information flow paths, thereby improving the effectiveness of any subsequent KD approach. Additionally, when there is a difference in the width of the teacher and student layers, we apply a pruning operation that reduces the width of the teacher's intermediate layers to align with the student's, enabling direct distillation.
The above strategy effectively enhances the performance of many baseline methods on a wide range of retrieval and classification experiments.

\section*{Acknowledgments}
This research was supported by the European Commission under the Horizon Europe project ELIAS (Grant Agreement 101120237), the H2020 projects AI4Media (Grant Agreement 951911) and MediaVerse (Grant Agreement 957252), and the Junior Star GACR project (Grant Agreement 21-28830M).

%%%%%%%%% REFERENCES
{\small
\balance
\bibliographystyle{ieee_fullname}
\bibliography{main}
}

\end{document}